\pdfoutput=1
\documentclass[11pt]{article}

\usepackage[final]{acl}

\usepackage{times}
\usepackage{latexsym}

\usepackage[T1]{fontenc}
\usepackage[utf8]{inputenc}

\usepackage{inconsolata}

\usepackage{hyperref}
\usepackage{url}
\usepackage{siunitx}
\usepackage{booktabs}       % professional-quality tables
\usepackage{amsfonts}       % blackboard math sy
\usepackage{xcolor}
\usepackage{nicefrac}       % compact symbols for 1/2, etc.
\usepackage[activate=true,
final,
tracking=false,   % don't space out smallcaps
kerning=true,
spacing=true,
factor=1050,
stretch=30,
shrink=30]{microtype}      % microtypography

\usepackage{enumitem}
\usepackage{pifont}   % For checkmark and cross

\usepackage{xspace}

\usepackage{url}
\usepackage{mathtools}
\usepackage{amsmath}
\usepackage{amsthm}
\usepackage{amssymb}
\usepackage{graphicx}
\usepackage{wrapfig}
\usepackage{subcaption}
\usepackage{bbm}
\usepackage{bm}
\usepackage{multirow}
\usepackage{booktabs}
\usepackage{xspace}

\usepackage[capitalize]{cleveref}
\crefname{page}{page}{pages}
\crefname{footnote}{footnote}{footnotes}   % "footnote" is lowercased, overriding capitalize option
\crefname{equation}{equation}{equations}   % "equation" is lowercased, overriding capitalize option; note that \labelcref drops this word if you want to say something like "the divergence (3)"
\crefname{corollary}{Corollary}{Corollaries}  % capitalized
\crefname{line}{line}{lines}               % "line" is lowercased, overriding capitalize option
\crefrangeformat{line}{lines #3#1#4--#5#2#6}
\crefname{lstlsting}{Listing}{Listings}   
\crefname{section}{\S}{\S\S}
\Crefname{section}{\S}{\S\S}    % must define start-of-sentence version explicitly since \S isn't a letter
\crefformat{section}{#2\S#1#3}  % remove space between section symbol and the number, and include \S in the hyperlink
\Crefformat{section}{#2\S#1#3}
\crefrangeformat{section}{\S\S#3#1#4--#5#2#6}
\Crefrangeformat{section}{\S\S#3#1#4--#5#2#6}
\crefmultiformat{section}{\S#2#1#3}{ and~\S#2#1#3}{, \S#2#1#3}{ and~\S#2#1#3}
\Crefmultiformat{section}{\S#2#1#3}{ and~\S#2#1#3}{, \S#2#1#3}{ and~\S#2#1#3}
\crefrangemultiformat{section}{\S\S#3#1#4--#5#2#6}{ and~\S\S#3#1#4--#5#2#6}{, \S\S#3#1#4--#5#2#6}{ and~\S\S#3#1#4--#5#2#6}
\Crefrangemultiformat{section}{\S\S#3#1#4--#5#2#6}{ and~\S\S#3#1#4--#5#2#6}{, \S\S#3#1#4--#5#2#6}{ and~\S\S#3#1#4--#5#2#6}
\usepackage{tcolorbox}
\usepackage{lipsum} % Optional, for dummy text
\usepackage{tabularx}
\usepackage{algorithm}  % the float environment
\usepackage[endLComment=]{algpseudocodex}

\definecolor{myDeepYellow}{rgb}{0.9412, 0.6902, 0.302}
\definecolor{myYellow}{rgb}{0.9765, 0.8824, 0.7255}
\definecolor{myBlue}{rgb}{0.6353, 0.7686, 0.8627}
\newcommand\better{\def\argii{}\docommandbetter}
\def\docommandbetter#1 {\colorbox{myBlue!80}{#1} \let\next\argii}
\def\argii{\let\next\docommandbetter}

\makeatletter
\newcommand{\smartperiod}{\@ifnextchar.{}{.\@\xspace}}
\newcommand{\smartcomma}{\@ifnextchar.{}{,}\xspace}
\makeatother
\newcommand{\latin}[1]{#1}  % or use \textit{#1} but that's rather precious
\newcommand{\eg}{\latin{e.g.}\smartcomma}
\newcommand{\ie}{\latin{i.e.}\smartcomma}

\newcommand{\spiritlm}{\textsc{SpiritLM}\xspace}
\newcommand{\wavvec}{\textsc{Wav2Vec2.0}\xspace}
\newcommand{\hubert}{\textsc{HuBERT}\xspace}
\newcommand{\whisper}{\textsc{Whisper}\xspace}

\newcommand{\llamatwo}{\textsc{LLaMa2}\xspace}
\newcommand{\llamathree}{\textsc{LLaMa3}\xspace}

\newcommand{\softalignment}{A^{\text{soft}}}
\newcommand{\unity}{\textsc{UnitY2}\xspace}
\newcommand{\charctc}{\textsc{Char-CTC}\xspace}
\newcommand{\subctc}{\textsc{Sub-CTC}\xspace}
\newcommand{\method}{\textsc{SSR-Connector}\xspace}

\title{SSR: Alignment-Aware Modality Connector for Speech Language Models}

\author{
  Weiting Tan$^{\spadesuit}$\thanks{\ \ Work was done during an internship at Meta AI.} \quad
  Hirofumi Inaguma$^{\heartsuit}$ \quad
  Ning Dong$^{\heartsuit}$\\
  \textbf{Paden Tomasello}$^{\heartsuit}$ \quad
  \textbf{Xutai Ma}$^{\heartsuit}$ \\
  $^{\spadesuit}$Johns Hopkins University \quad
  $^{\heartsuit}$Meta AI Research
}

\begin{document}
\maketitle
\begin{abstract}

Fusing speech into a pre-trained language model (SpeechLM) usually suffers from the inefficient encoding of long-form speech and catastrophic forgetting of pre-trained text modality. We propose \method (Segmented Speech Representation Connector) for better modality fusion. Leveraging speech-text alignments, our approach segments and compresses speech features to match the granularity of text embeddings. Additionally, we introduce a two-stage training pipeline that includes the distillation and fine-tuning phases to mitigate catastrophic forgetting. \method outperforms existing mechanism for speech-text modality fusion, consistently achieving better speech understanding (\eg $+10$ accuracy on StoryCloze and $+20$ on Speech-MMLU) while preserving pre-trained text ability.

\end{abstract}

\section{Introduction}
% brief intro of LLM and MLLM
Large language models \citep[LLMs]{gpt3,palm,vicuna2023,palm2,touvron2023llama,openai2024gpt4, grattafiori2024llama3herdmodels, deepseekai2025deepseekr1incentivizingreasoningcapability} have demonstrated remarkable performance across various tasks and extending pre-trained abilities from LLMs to other modalities has sparked interest in multimodal LLMs \citep{alayrac2022flamingo,liu2023visual,openai2024gpt4,tang2024salmonn,kyutai2024moshi}. In this work, we focus on integrating speech into pre-trained language models (SpeechLMs). A straightforward approach is to transcribe speech into text and use these transcriptions as prompts for large language models \citep{audiogpt}; however, such cascaded systems suffer from error propagation, higher latency, and cannot leverage raw speech information like emotion, speaker identity, and other paralinguistic cues \citep{faruqui2021revisitingboundaryasrnlu, Lin_2022, kim2024integratingparalinguisticsspeechempoweredlarge}. Consequently, developing end-to-end SpeechLMs that directly fuse speech or audio input has gained popularity, where various approaches have been explored to encode speech and align its representation with pre-trained language models \citep{speechgpt,rubenstein2023audiopalm,yu2023connectingspeechencoderlarge,voxtlm,hassid2024textually,tang2024salmonn,nguyen2024spiritlm}.

\begin{figure}[t]
    \centering
    \includegraphics[scale=0.65]{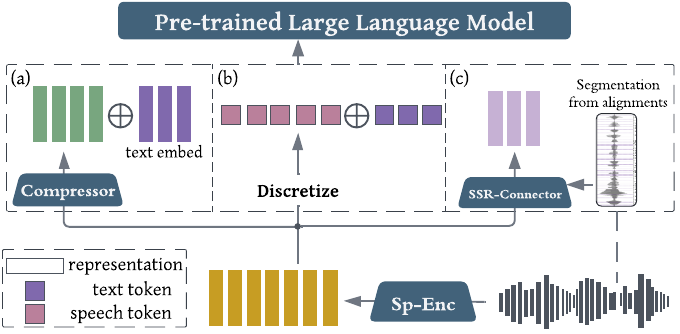}
    \caption{Comparison of different approaches for speech-text modality fusion. \textit{(a)}: compressor-based connector. \textit{(b)}: direct fusion with speech units. \textit{(c)}: our alignment-aware connector.}
    \label{figure::overview}
    \vspace{-1em}
\end{figure}

Speech representations can be integrated into pre-trained language models mainly through two approaches. The first method involves using connector modules that align speech representations with the language model’s input space without modifying the model’s existing vocabulary. These connector-based techniques typically incorporate a compression module to shorten the speech features, enhancing efficiency. However, connectors are generally first trained for the speech recognition task (with concatenated speech-to-text data) and \textbf{lack the ability to support text or speech generation unless further instruction-finetuned}. 

The second approach, unit-based fusion, directly incorporates discrete speech units—normally derived from self-supervised models like HuBERT \citep{hsu2021hubert}, XLS-R \citep{xlsr}, or DinoSR \citep{dinosr}—into the language model’s vocabulary. This allows the language model to be fine-tuned with a combination of speech and text tokens, enabling it to handle dual-modal inputs and outputs. Despite its versatility, \textbf{unit-based fusion can lead to longer and less efficient training contexts} due to the sparser nature of speech information. Regardless of the fusion approach, SpeechLMs often face the challenge of catastrophic forgetting, where the model loses its pre-trained text capabilities \citep{tang2024salmonn, nguyen2024spiritlm, kyutai2024moshi}.

To tackle these challenges, we propose \method (Segmented Speech Representation Connector), which grounds speech representations in the same semantic space as transcription token embeddings. Different from prior work that concatenates speech with text (\cref{figure::overview} (a,b)) for modality fusion, we leverage speech-text alignments to segment and compress speech features (\cref{figure::overview} (c)).

To mitigate catastrophic forgetting when introducing the speech modality, we propose a two-stage training pipeline. In Stage 1, we freeze the LLM and pre-train the connector using speech-text distillation, adapting speech inputs into compressed representations semantically aligned with text embeddings. In Stage 2, we unfreeze the LLM and fine-tune it using next-token prediction, with the adapted representation as input and the corresponding transcription tokens as targets.

\method outperforms prior SpeechLMs, including \spiritlm, \textsc{Voxtlm}, \textsc{TWIST}, and \textsc{AudioLM} \citep{nguyen2024spiritlm, voxtlm, hassid2024textually, borsos2023audiolm}, across multiple tasks. These include Prompt-based Automatic Speech Recognition (ASR) and Spoken Language Understanding with sWUGGY, sBLIMP, and StoryCloze \citep{nguyen2020zeroresourcespeechbenchmark, storycloze_dataset}. Our approach also improves performance on Massive Multitask Language Understanding (MMLU) \citep{mmlu_dataset} and its speech-based counterpart, Speech-MMLU, which we introduce to assess cross-modal reasoning. Finally, we analyze different training strategies (\cref{sec::stage2}) and speech-text aligners (\cref{sec::aligner}) for \method.

\section{Related Work}
\paragraph{Modality Fusion for Speech Language Models} 
SpeechLM typically encodes audio waveforms into high-dimensional features using pre-trained encoders and integrate these representations to pre-trained LLMs via a connection (adapter) module \citep{asr_llama, yu2023connectingspeechencoderlarge, speechgpt, tang2024salmonn}. To compress speech representations, \citet{fathullah2023promptinglargelanguagemodels} apply stacking-based fixed-rate compression on speech features extracted from the Conformer model \citep{conformer}. Inspired by the Q-former architecture \citep{blip2}, \citet{yu2023connectingspeechencoderlarge} compress speech features using a fixed number of query tokens, while \citet{tang2024salmonn} extend this approach to a window-level Q-former to support variable frame-rate reduction. Alternatively, \citet{asr_llama} utilize Connectionist Temporal Classification (CTC) \citep{ctc} to compress representations.

Besides connector-based modality fusion, pre-processing other modalities—such as speech, vision, and videos—into tokens \citep{macaw_llm, li2023llamavidimageworth2, chameleonteam2024chameleonmixedmodalearlyfusionfoundation, kondratyuk2024videopoetlargelanguagemodel} has attracted attention for its scalability. Speech units are typically extracted from self-supervised representations. For instance, AudioLM \citep{borsos2023audiolm} integrates semantic tokens from w2v-BERT \citep{w2vbert} and acoustic tokens from SoundStream \citep{zeghidour2021soundstream} for autoregressive audio generation. \citet{rubenstein2023audiopalm} fine-tune the pre-trained LLM PaLM-2 \citep{palm2} with audio tokens processed by AudioLM, enabling both text and speech as input and output. Similarly, VoxtLM \citep{voxtlm} performs multi-task training with speech units and text tokens, achieving high-quality speech recognition and synthesis. To mitigate catastrophic forgetting, \citet{nguyen2024spiritlm} propose an interleaved training mechanism to fuse speech tokens into \llamatwo model \cite{touvron2023llama}.

\begin{figure*}[t]
    \vspace{-1em}
    \centering
    \includegraphics[scale=1]{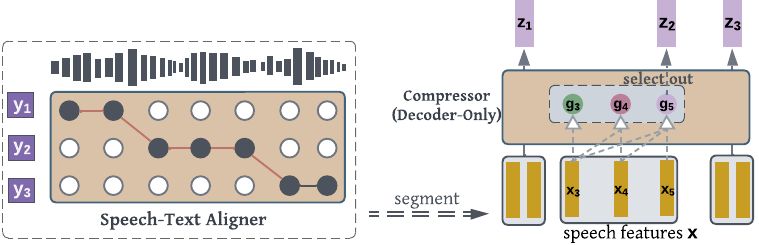}
    \vspace{-0.5em}
    \caption{\method compresses speech features using speech-text alignments. Features are transformed by a Decoder-only model and selected at boundary index of each segment.}
    % \vspace{-1em}
    \label{figure::adapter_module}
\end{figure*}

\noindent\textbf{Speech-text Alignment Extraction} 
Various aligner tools are available for extracting speech-text alignments. For example, the Montreal Forced Aligner \citep[MFA]{McAuliffe2017MontrealFA} is an easy-to-use tool based on the Kaldi toolkit \citep{povey2011kaldi}. Connectionist Temporal Classification (CTC) \citep{ctc} is also widely used for speech-text alignment \citep{Sainath2020EmittingWT, huang2024peakyaccuratectcforced}; since it is a by-product of speech recognition, it supports alignment without explicit text labels. More recently, the UnitY2 aligner \citep{seamless} and the ZMM-TTS aligner \citep{gong2024zmmttszeroshotmultilingualmultispeaker} have shown excellent alignment performance across multiple languages. These aligners rely on speech units extracted from pre-trained encoders \citep{w2v2, hsu2021hubert, xlsr} and use variants of RAD-TTS \citep{shih2021radtts} as their alignment backbone.
 % like HuBERT \citep{hsu2021hubert}, XLS-R \citep{xlsr}, and Wav2Vec2.0 \citep{w2v2},

\section{Methodology}\label{sec::method}
We develop an alignment-aware speech representation connector to foster modality fusion between speech and pre-trained language model. We introduce our connector design in \cref{sec::adapter_overview} and present our two-stage training pipeline in \cref{sec::training_method}.

\subsection{Alignment-Aware Speech Representation Connector}\label{sec::adapter_overview}
% overall method: obsure the detail of how alignment is extracted. Only present the high level design and training objective for distillation and fine-tuning
Though previous connectors \citep{fathullah2023promptinglargelanguagemodels, yu2023connectingspeechencoderlarge, asr_llama, tang2024salmonn} vary in their compressor designs, they do not explicitly leverage speech-text alignment information. \method, in contrast, uses speech-text alignments to segment and compress speech features into the same granularity as text tokens. As illustrated in \cref{figure::adapter_module}, our connector consists of two components: (1) a speech-text aligner and (2) a feature compressor.

Given speech features $\bm x = (x_1, \cdots, x_n) \in \mathbb{R}^{n\times D}$ extracted by pre-trained speech encoders (e.g., \wavvec, \hubert, \whisper, etc.), the aligner produces a monotonic mapping (alignment path) between the speech features and their transcriptions $\bm y = (y_1, \cdots, y_m) \in \mathbb{R}^{m \times 1}$. This mapping can be computed based on both speech features (or their units) and transcriptions \citep{seamless, gong2024zmmttszeroshotmultilingualmultispeaker}, or solely based on speech input \citep{Sainath2020EmittingWT, dong2020cif, huang2024peakyaccuratectcforced}. We abstract away the aligner's implementation here but provide detailed description and comparison of various aligners in \cref{sec::aligner}. 

Using the alignment mapping, we segment the input into $m$ chunks of speech features, where each chunk semantically corresponds to a transcription token. For example, in \cref{figure::adapter_module}, speech features are segmented at indices $(2, 5, 7)$ according to the alignment path. We refer to these indices as boundary indices. Once the boundary indices are identified, we first apply a linear layer to transform the speech features to match the embedding dimension $H (H > D)$ of the pre-trained LLM, since LLMs typically have a larger feature dimension than pre-trained speech encoders. We then use the boundary indices to aggregate and compress the speech representations in each chunk through a Transformer Decoder model \citep{transformer}.

Specifically, we apply a causal decoder-only model to transform speech features into high-dimensional representations $\bm g = f(\bm x; \theta_{\text{dec}}) \in \mathbb{R}^{n\times H}$. Since each position incorporates past context, we adopt a selection-based compression method \cite{tan2024streaming}, using boundary-indexed features from $\bm g$ to form the compressed representation $\bm z \in \mathbb{R}^{m\times H}$. While our initial design used a block-wise attention mask to limit cross-chunk information flow (as shown in \cref{figure::adapter_module}), we found that removing these masks simplifies training and inference with minimal performance loss (\cref{sec::main_results}).

\subsection{Training Method}\label{sec::training_method}
% describe stage 1 and 2 training in an abstract way
Previous approaches to integrate speech into LLMs typically use speech-text data concatenated in ASR format (\ie speech representation followed by its transcription text embedding), to pre-train the connector \citep{yu2023connectingspeechencoderlarge, asr_llama, tang2024salmonn}. However, after such pre-training, the model is limited to speech recognition task and necessitates another instruction-tuning stage to perform generative tasks with pre-trained connectors \citep{speechgpt, tang2024salmonn}. Moreover, once the LLM is unfrozen and fine-tuned (whether based on a pre-trained connector or direct fusion with speech units), it suffers from catastrophic forgetting, leading to degraded text capabilities \citep{nguyen2024spiritlm, tang2024salmonn}. 

\begin{figure}[t]
    % \vspace{-1em}
    % \hspace{4em}
    \centering
    \includegraphics[scale=0.77]{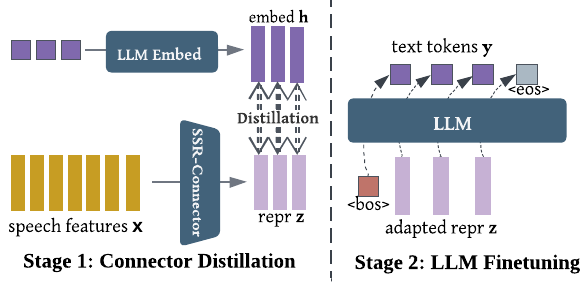}
    \vspace{-2em}
    \caption{Two-stage training pipeline for SpeechLM with our alignment-aware modality connector.}
    \vspace{-1em}
    \label{figure::train_method}
\end{figure}

With \method, we convert speech into representations with the same granularity as their transcription tokens. This allows us to fine-tune the SpeechLM directly using the next-token prediction objective, where the input is the compressed representation $\bm z$ and the target is the transcription $\bm y$. This approach is possible because our feature $\bm z$ and text token $\bm y$ share the same length $m$. \textbf{However, our preliminary studies showed that directly fine-tuning with the next-token prediction objective leads to catastrophic forgetting, undermining the pre-trained LLM's abilities}. Therefore, we propose a two-stage training pipeline consisting of a distillation stage and a fine-tuning stage (visualized in \cref{figure::train_method}).

In Stage 1, we pre-train \method by distilling the LLM's text embeddings to align the connector's representations with the LLM's embedding space. Formally, given aligned speech-text data, we can compute the text embeddings $\bm h = f(\bm y; \theta_{\text{emb}})$, where $\bm y$ is the transcription token sequence, $\theta_{\text{emb}}$ is the embedding table, and $f$ maps tokens $\bm y$ to their embeddings. Following our connector design in \cref{sec::adapter_overview}, we then obtain the compressed speech representations $\bm z$. For distillation, we use a combination of cosine similarity loss $\mathcal{L}_{\text{cos}}$ and mean squared error (MSE) loss $\mathcal{L}_{\text{MSE}}$
\begin{equation} 
\begin{aligned}
\mathcal{L} &= \lambda \mathcal{L}{\text{cos}} + \mathcal{L}_{\text{MSE}} \\
&=\frac{1}{m}\sum_{i=1}^{m} \left[ \lambda \left(1 - \frac{\mathbf{z}_i^\top \mathbf{h}_i}{|\mathbf{z}_i|\cdot|\mathbf{h}_i|} \right) + \left| \mathbf{z}_i - \mathbf{h}_i \right|^2 \right]
\end{aligned}
\end{equation}
where $\lambda$ is a hyperparameter to balance the losses\footnote{In practice, we set $\lambda=5$ to balance the scales of the cosine similarity and MSE losses}. In Stage 2, we fine-tune the LLM with the pre-trained speech connector using the next-token prediction objective. We freeze the speech connector and update only the LLM's parameters using the negative log-likelihood (NLL) loss:

\begin{equation}
\mathcal{L}_{\text{NLL}} = -\sum_{t=1}^m \log p(y_t \mid \bm z_{<t}; \theta_{\text{LLM}}) \end{equation}
where $y_t$ is the $t^{\text{th}}$ token in the transcription sequence $\bm y$, $\bm z_{<t}$ denotes all preceding speech representations, and $\theta_{\text{LLM}}$ represents the LLM's parameters. Note that our NLL loss is computed using only the preceding speech representations $\bm z_{<t}$ (see \cref{figure::train_method}), whereas previous methods \citep{asr_llama, tang2024salmonn} condition on both speech information and preceding text tokens $\bm y_{<t}$.

In \cref{sec::stage1}, We demonstrate the performance of SpeechLM after distillation training. In \cref{sec::stage2}, we present results after fine-tuning SpeechLM and compare various fine-tuning strategies to identify the method that minimizes catastrophic forgetting.

\section{Stage 1: Alignment-Aware Connector Distillation}\label{sec::stage1}
\subsection{Datasets}\label{sec::dataset}
For distillation training, we use the aligned speech-to-text dataset MLS \citep{mls_dataset}, specifically the English portion, which consists of about 50,000 hours of speech. To evaluate our SpeechLMs, we employ different benchmark datasets (see \cref{table::datasets_main}). To assess the model's spoken language understanding (SLU) capabilities, we follow \citet{nguyen2024spiritlm} and use sWUGGY, sBLIMP, and the StoryCloze dataset. sWUGGY evaluates whether a model can discriminate between real spoken words and non-words (e.g., ``brick" vs. ``blick"), while sBLIMP assesses if the model can distinguish between a grammatically correct spoken sentence and its ungrammatical variant. We evaluate our SpeechLMs on both text ($T$) and speech ($S$) versions of sWUGGY and sBLIMP. 

The StoryCloze dataset measures whether the model can identify the plausible ending between two sentences given the beginning of a short story, which typically requires high-level semantic understanding and common sense \citep{storycloze_dataset}. Besides spoken and text versions of StoryCloze, following \citet{nguyen2024spiritlm}, we use a speech-text version ($S\rightarrow T$), where the beginning of the story is synthesized into speech and the two ending sentences are kept in text format. This version requires the model to have cross-modal understanding to infer the sensible story ending.

\begin{table*}[t]
\centering
\small
\begin{tabular}{l l l l}
\toprule
\textbf{Eval Dataset} & \textbf{Type} & \textbf{Eval Metric} &\textbf{Eval Modality}\\
\midrule
sWUGGY \citep{nguyen2020zeroresourcespeechbenchmark} & Choice Task     & Accuracy   &$S, T$      \\
sBLIMP \citep{nguyen2020zeroresourcespeechbenchmark} & Choice Task     & Accuracy   &$S, T$      \\
StoryCloze \citep{storycloze_dataset}                & Choice Task     & Accuracy   &$S, T$, $S\rightarrow T$      \\
MMLU \citep{mmlu_dataset}                            & Choice Task     & Accuracy    &$T$     \\
Speech-MMLU (\textit{Ours})                                          & Choice Task     & Accuracy    &$S\rightarrow T$     \\
LibriSpeech \citep{librisppech_dataset}          & Generation Task & Word Error Rate &$S\rightarrow T$  \\
\bottomrule
\end{tabular}
\vspace{-0.5em}
\caption{Evaluation Datasets and their types. For the evaluation format, $S$ is speech-only, $T$ is text-only, and $S\rightarrow T$ means the evaluation prompt consists of speech prefix and text continuation.}
\vspace{-1em}
\label{table::datasets_main}
\end{table*}

MMLU \citep{mmlu_dataset} is widely used to assess LLMs' knowledge comprehension, understanding, and reasoning abilities, and we use it to measure the extent of forgetting during cross-modal fine-tuning. Since MMLU is a diverse and high-quality evaluation dataset for LLMs, we craft a variant, Speech-MMLU, to assess our SpeechLM's cross-modal understanding. Specifically, we utilized \textsc{AudioBox} \citep{vyas2023audioboxunifiedaudiogeneration}, a high-quality text-to-speech synthesizer, to convert the question portion of each choice task into speech while keeping the multiple-choice answers in text format. We selected a subset of MMLU to construct our Speech-MMLU dataset, as some domains' questions are not suitable for synthesis (\eg the algebra subset contains many mathematical notations that are not synthesized properly).

sWUGGY, sBLIMP, StoryCloze, and Speech-MMLU are all categorized as "Choice Task", meaning several choices are presented to the SpeechLM (Speech-MMLU has four choices while the other task has only two choices). For each task, we compute accuracy using groundtruth choice and the highest likelihood choice predicted by the SpeechLM. Lastly, we also evaluate our SpeechLM's ASR performance using the Librispeech clean/other datasets. We evaluate ASR in a prompt-based fashion with zero-shot and five-shot setting. Comprehensive details about our datasets and evaluation can be found in \cref{app::dataset_details}.

\subsection{Model Setup}\label{sec::model_setup}
We instantiate our LLM using the pre-trained \llamathree model \citep{grattafiori2024llama3herdmodels} and employ DinoSR \citep{dinosr} as our pre-trained speech feature extractor. Our speech connector includes a linear layer that maps DinoSR's extracted representations ($D=768$) to the LLM's embedding space dimension ($H=4096$). We then utilize a 4-layer Transformer Decoder to transform and compress the speech representations based on alignments, as described in \cref{sec::adapter_overview}. The compressed representations $\bm z$ and the embeddings of text tokens $\bm h$ are used to compute the distillation loss for updating the connector's parameters. We train our connector for 400,000 steps with a learning rate of $1 \times 10^{-5}$, using dynamic batching with a maximum of 4,096 tokens per device. We employ distributed data parallelism (DDP) with 32 A100 GPUs.

To extract alignments, we experimented with various approaches, including the \unity aligner, CTC-based aligners \cite{ctc}, and Continuous Integrate-and-Fire \citep[CIF]{dong2020cif}. Due to space constraints, we provide comprehensive descriptions and comparisons of these methods in \cref{sec::aligner}, where we evaluate both the alignment quality and the Word Boundary Error of the segmentations. After assessing their performance, we selected \unity \cite{communication2023seamless} and character-level CTC (\charctc) as our connector backbone to report experimental results. Overall, \unity offers superior alignment quality because it utilizes both speech and text as input. In contrast, CTC only requires speech input to compute segmentation for our connector.

\begin{table*}[t]
\centering
\small
\begin{tabular}{@{}lcccccccc@{}}
\toprule
\textbf{Model Type} & \multicolumn{2}{c}{\textbf{sWUGGY}} & \multicolumn{2}{c}{\textbf{sBLIMP}} & \multicolumn{3}{c}{\textbf{Storycloze}} & \textbf{MMLU} \\ 
\cmidrule(lr){2-3} \cmidrule(lr){4-5} \cmidrule(lr){6-8} \cmidrule(l){9-9}
                    & T             & S                   & T             & S                   & T             & S                   & S$\rightarrow$T             & 5-shot             \\ 
\midrule
\textit{\quad Previous Work} \\
\textsc{GSLM}${}^{\diamondsuit}$ \citep{gslm}& $\emptyset$ & 64.8 & $\emptyset$ & 54.2  & $\emptyset$ & 53.3  & $\emptyset$  & $\emptyset$ \\
\textsc{AudioLM}${}^{\diamondsuit}$ \citep{borsos2023audiolm}& $\emptyset$ & 71.5 & $\emptyset$ & 64.7  & $\emptyset$ & \underline{\ \ }  & $\emptyset$  & $\emptyset$ \\
\textsc{Voxtlm}${}^{\diamondsuit}$ \citep{voxtlm}& \better 80.3 & 66.1 & \better 74.2 & 57.1  & \underline{\ \ } & \underline{\ \ }  & \underline{\ \ }  & \underline{\ \ } \\
\textsc{TWIST}${}^{\diamondsuit}$ \citep{hassid2024textually}& $\emptyset$ & \textbf{74.5} & $\emptyset$ & 59.2  & $\emptyset$ & 55.4  & $\emptyset$ & $\emptyset$ \\
\textsc{Moshi}${}^{\clubsuit}$ \citep{kyutai2024moshi}                   & $\emptyset$ &  72.6  & $\emptyset$ & 58.8 & $\emptyset$ & 60.8   & \underline{\ \ }  & 49.8  \\
\textsc{SpiritLM}${}^{\diamondsuit}$ \citep{nguyen2024spiritlm}                   & \better 80.3 & 69  & 73.3  & 58.3 & \better 79.4 & 61   & 64.6  & 36.9   \\
\spiritlm (\llamathree)${}^{\spadesuit}$                   & 77.6 & \better 73.5 & \textbf{74.5} & 56.3 & 75.1 & 61.1 & 61.6  & \better 53.5 \\ 

\midrule
\quad \method\\
\unity + Blockwise-mask & \textbf{81} & 71.5 & \textbf{74.5} & \textbf{73.1}  & \textbf{80.9} & \textbf{71.8}  & \textbf{75}  & \textbf{65.3} \\
\unity   & 81 & 71.2 & 74.5 & \better 72.4  & 80.9 & \better 69.3 & \better 74.8  &  65.3     \\
\charctc & 81 & 56.4 & 74.5 & 67.3  & 80.9 & 62.2  & 74.3 &  65.3 \\
\charctc (Unit-based) & 81 & 54.1 & 74.5 & 61.8 & 80.9 & 59.2 & 72.5 & 65.3 \\
% \midrule
% \textit{\quad Ours Fine-tuned}\\
% \charctc & 82.5 & 56.6 & 75.8 & 68.8 & 75.2 & 62.8 & 71 & 57.4\\
% \charctc \textit{with} text-only & 82.9 & 56.7 & 75.9 & 68.9 & 81 & 63.4 & 73.1 & 63.1\\
\midrule
\textit{\quad Cascade System}\\
ASR (\whisper) + \llamatwo${}^{\diamondsuit}$ & 84.1 & 79.2 & 72.8 & 71.6  & 81.9 & 75.7  & 75.7 &  46.2 \\

\bottomrule
\end{tabular}
\caption{Model performance (accuracy) on spoken language understanding and MMLU. ${}^{\diamondsuit}$: Results taken from \citet{nguyen2024spiritlm}.${}^{\clubsuit}$: Results taken from \citet{kyutai2024moshi}.  ${}^{\spadesuit}$: Our implementation of \spiritlm based on \llamathree checkpoint. We fill with $\emptyset$ the task and modality that are not supported by the reported system, and with \underline{\ \ } the scores that are not publicly available. We bold the best result and highlight the second-best system with the blue color box (excluding the cascaded system).}
\vspace{-1em}
\label{table::main_results}
\end{table*}

\subsection{Experimental Results}\label{sec::main_results}

In this section, we present the evaluation of \method based SpeechLM in terms of Spoken Language Understanding (SLU) and Cross-modal Understanding (through our use of Storycloze and Speech MMLU benchmark). We also evaluate our model with prompting-based speech recognition and speech style recognition.

We compare against several systems that varies in training approaches (pre-trained from scratch or fine-tuned), types of speech units, and the size of training data. Briefly, GSLM \citep{gslm} trains on speech units like HuBERT, TWIST \citep{hassid2024textually} is a textually pretrained speech model based on Llama-13B \citep{touvron2023llama}, and AudioLM \citep{borsos2023audiolm} employs a cascade system with a semantic sequence model alongside coarse- and fine-acoustic models. These models focus solely on speech without capabilities for text understanding or generation. More recently, \spiritlm \citep{nguyen2024spiritlm} and VoxtLM \citep{voxtlm} have adopted multi-task training objectives that incorporate text-only, speech-only, and speech-text token sequences to fuse the speech modality into pre-trained LLMs effectively. Since the original \spiritlm is fine-tuned based on \llamatwo, we follow the same recipe to fine-tune the \llamathree-based \spiritlm ourselves for a fair comparison on text-relevant metrics like MMLU.

\paragraph{Spoken Language Understanding Performance} As shown in \cref{table::main_results}, our systems outperform previous models on all tasks except sWUGGY. The sWUGGY dataset includes incorrectly spoken words that cause segmentation errors because these words were not present during aligner training, leading to our system's lower performance on this dataset. However, sWUGGY is the least significant task since it relies on synthesized incorrect words and does not require the model’s understanding or reasoning capabilities. In contrast, both \unity and \charctc based connector greatly surpass previous models on other datasets, demonstrating the effectiveness of \method in enhancing SLU performance while preserving model's text understanding ability.

Beyond \unity and \charctc, we introduce two additional systems for ablation. The \textbf{\unity + Blockwise-mask} system achieves the highest performance by applying a blockwise attention mask to further constrain the Transformer-Decoder (described in \cref{sec::adapter_overview}). However, due to its marginal improvement over \unity and increased computational cost, we decide to simplify the design and remove the blockwise-attention masks. The \textbf{\charctc (Unit-based)} system differs by utilizing discrete speech units instead of raw waveform features processed by the DinoSR \citep{dinosr} encoder. These units are extracted via K-Means clustering on DinoSR representations, which leads to some information loss during discretization and reconstruction, resulting in lower performance compared to \charctc. Nonetheless, \charctc (Unit-based) demonstrates that \textit{our alignment-aware connector design is compatible with discrete speech units as well}.

\begin{table*}[t]
\centering
\small
\begin{tabular}{lcc|cc|cc}
\toprule
\textbf{Model Type}  & \multicolumn{2}{c|}{\textbf{Speech MMLU} $\uparrow$ } & \multicolumn{2}{c|}{\textbf{ASR Clean Test} $\downarrow$} & \multicolumn{2}{c}{\textbf{ASR Other Test} $\downarrow$} \\
\cmidrule(lr){2-3} \cmidrule(lr){4-5} \cmidrule(lr){6-7}
& \textbf{0-shot} & \textbf{5-shot} & \textbf{0-shot} & \textbf{5-shot} & \textbf{0-shot} & \textbf{5-shot} \\
\midrule
\spiritlm \citep{nguyen2024spiritlm}     & N/A   & N/A    & N/A    & 21.9$^*$ & N/A    & 29.2$^*$ \\
\spiritlm (\llamathree)   & 40.5  & 42.75 & N/A    & 21.0$^*$ & N/A   & 28.5$^*$ \\
\midrule
\quad \method\\
\unity + Blockwise-mask   & \textbf{65.0} & \textbf{69.5} & \textbf{5.0}  & \textbf{2.6}     & \textbf{8.1}  & \textbf{6.8} \\
\unity                    & 64.2 & 68.6 & 5.6   & 4.0     & 12.1 & 10.6 \\
\charctc                  & 61.7 & 66.5 & 9.7  & 6.5     & 20.2 & 14.9 \\
\charctc (Unit-based)     & 57.4 & 62.3 & 12.6 & 8.8     & 25.6 & 18.6 \\
\bottomrule
\end{tabular}
\vspace{-0.5em}
\caption{Comparison of Speech-MMLU and ASR performance. Speech-MMLU results are micro-averages across all domains. $^*$: For \spiritlm, We report WER using 10-shot prompting, following \citet{nguyen2024spiritlm}.}
\vspace{-1em}
\label{table::speech_mmlu_asr}
\end{table*}

\paragraph{Speech-MMLU and Prompt-based ASR Performance} In addition to SLU tasks, we evaluate our systems on the Speech-MMLU benchmark, which assesses cross-modal understanding and is more challenging than previous SLU tasks. We also conduct prompt-based ASR evaluations to assess the quality of the adapted features. As shown in \cref{table::speech_mmlu_asr}, our systems greatly outperform the previous SpeechLM (\spiritlm), achieving a +20 accuracy improvement on the Speech-MMLU dataset\footnote{~We report micro-average across 22 domains and the detailed breakdown is available in \cref{app::speech_mmlu}.}. These results indicate that SpeechLM based on \method possesses enhanced cross-modal abilities that enable it to comprehend spoken questions and reason through multiple-choice options to select correct answers. Similarly, our systems achieve much lower WERs on the Librispeech clean and other test sets compared to \spiritlm. Notably, neither \spiritlm nor our system was trained on ASR tasks, \textit{so the model relies solely on in-context learning to generate transcriptions}. 

We also compared our system against another connector-based system, \textsc{salmonn} \citep{tang2024salmonn}, over Storycloze and Speech MMLU (both in $S\rightarrow T$ format) and we find that \textsc{salmonn} achieved an accuracy of $63.3\%$ on Storycloze and $25.3\%$ on Speech-MMLU, while our system has over $74\%$ accuracy on Storycloze and over $60\%$ accuracy on Speech-MMLU. The result indicates that catastrophic forgetting remains a severe issue for previous connector-based methods as well.

\paragraph{Beyond Semantics}
In \cref{table::style_recognition}, we also show that the connector retains paralinguistic information. We evaluate this using the Expresso benhmark \cite{expresso_benchmark} by prompting our model to predict speech styles. Our SpeechLM can distinguish expressions through in-context learning without being fine-tuned for emotion recognition (we also provide the cascaded baseline (Whisper + \llamathree) as a baseline where style can only be inferred from transcriptions). More experimental details are provided in \cref{app::paralinguistic}. This analysis demonstrates that our connector preserves non-semantic information even though we focus on aligning semantics and reducing catastrophic forgetting. Our connector design also complements existing methods for emotion recognition, such as using expressive tokens in SpiritLM \cite{nguyen2024spiritlm} and emotion-relevant instruction tuning in SALMONN \cite{tang2024salmonn}.

\begin{table}[t]
    \centering
    \small
    \resizebox{\linewidth}{!}{
    \begin{tabular}{lcccc}
        \toprule
        \textbf{Task} & \textbf{Model} & \textbf{0-shot} & \textbf{5-shot} & \textbf{10-shot} \\
        \midrule
        \multirow{2}{*}{Whisper vs. Laugh} & Cascaded & 51.6 & 52.2 & 54.7 \\
                                         & Ours     & 49.6 & 64.0 & 75.9 \\
        \midrule
        \multirow{2}{*}{Happy vs. Sad}     & Cascaded & 50.0 & 51.8 & 51.0 \\
                                         & Ours     & 51.6 & 52.2 & 54.7 \\
        \bottomrule
    \end{tabular}}
    \vspace{-1em}
    \caption{Accuracy of Speech Style Recognition with In-context Learning}
    \label{table::style_recognition}
    \vspace{-1em}
\end{table}

\begin{figure*}[h]
    \centering
    \vspace{-1em}
    \includegraphics[scale=0.45]{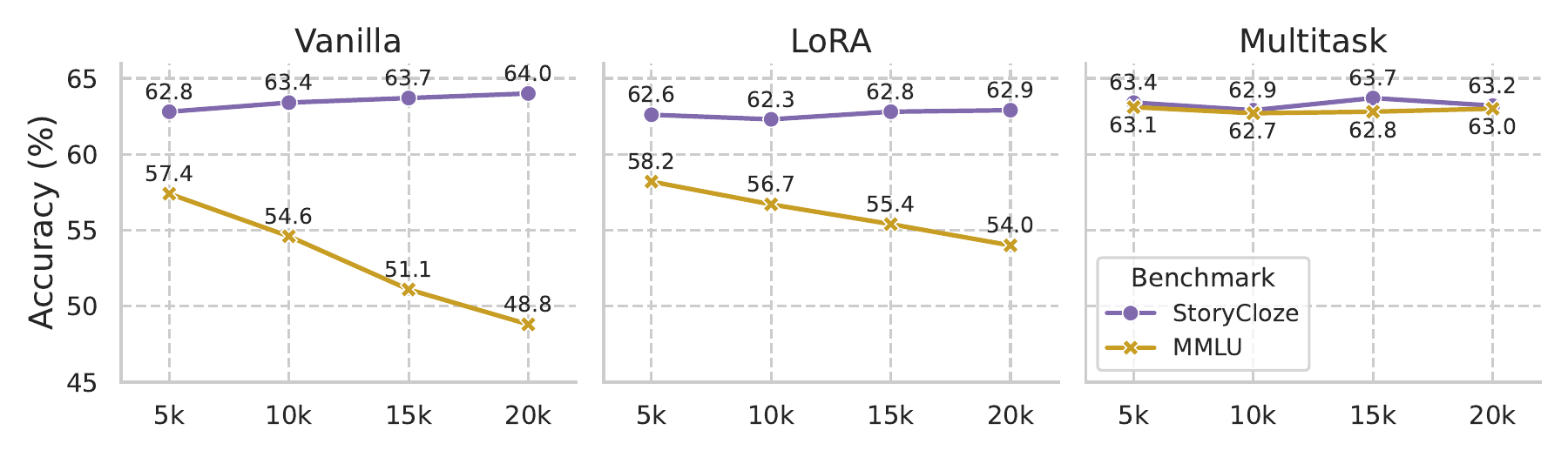}
    \vspace{-1.5em}
    \caption{Comparison of different fine-tuning methods on StoryCloze ($S$) and MMLU benchmark.}
    \label{figure::trade_off}
    % \vspace{-1em}
\end{figure*}

\begin{table*}[t]
\centering
\small
\resizebox{\linewidth}{!}{
\begin{tabular}{l|cc|cc|ccc|c|cc|cc}
\toprule
\textbf{Model Type} & \multicolumn{2}{c|}{\textbf{sWUGGY}} & \multicolumn{2}{c|}{\textbf{sBLIMP}} & \multicolumn{3}{c|}{\textbf{Storycloze}} & \multicolumn{1}{c|}{\textbf{MMLU}} & \multicolumn{2}{c|}{\textbf{Speech MMLU}} & \multicolumn{2}{c}{\textbf{ASR (5-shot)}$\,\downarrow$} \\ 
\cmidrule(lr){2-3} \cmidrule(lr){4-5} \cmidrule(lr){6-8} \cmidrule(lr){9-9} \cmidrule(lr){10-11} \cmidrule(lr){12-13}
 & T & S & T & S & T & S & S$\rightarrow$T & 5-shot & 0-shot & 5-shot & Clean & Other \\ 
\midrule
\spiritlm (\llamathree)   & 77.6 & 73.5 & 74.5 & 56.3 & 75.1 & 61.1 & 61.6 & 53.5 & 40.5 & 42.8 & 21.0$^*$ & 28.5$^*$ \\
\midrule
\charctc                  & 81.0 & 56.4 & 74.5 & 67.3 & 80.9 & 62.2 & \textbf{74.3} & \textbf{65.3} & \textbf{61.7} & \textbf{66.5} & 6.5 & 14.9  \\
+ Multitask Finetuning            & \textbf{82.9} & \textbf{56.7} & \textbf{75.9} & \textbf{68.9} & \textbf{81.0} & \textbf{63.4} & 73.1 & 63.1 & 48.1 & 56.3 & \textbf{5.7} & \textbf{13.1} \\
\bottomrule
\end{tabular}
}
\vspace{-1em}
\caption{
Performance comparison when the model is fine-tuned. $^*$: For \spiritlm, WER is reported using 10-shot prompting for ASR, following \citet{nguyen2024spiritlm}. We observe that stage 2 fine-tuning enhances the model’s performance on speech-only tasks but compromises its cross-modal capabilities.
}
\vspace{-1em}
\label{table::finetune_results}
\end{table*}

\section{Stage 2: Speech Language Model Fine-tuning}\label{sec::stage2}
In Stage 1 (\cref{sec::stage1}), we freeze the pre-trained LLM and distill its text embeddings into our alignment-aware connector. In this section, we fine-tune SpeechLM by freezing the connector and updating the LLM. This process enhances the model's spoken language understanding (SLU) performance by fitting SpeechLM on the aligned speech-text data, albeit at the expense of degrading its pre-trained text capabilities. In the following sections, we compare various methods to mitigate catastrophic forgetting and demonstrate their trade-offs between speech and text understanding.

\subsection{Mitigate Catastrophic Forgetting}\label{sec::mitigate_forgetting}
\paragraph{Model and Dataset Setup} We fine-tune SpeechLM using the next-token prediction objective described in \cref{sec::training_method}. In this stage, we freeze the connector distilled in Stage 1 and unfreeze the LLM (\llamathree) parameters. Following Stage 1 (\cref{sec::stage1}), we use the MLS dataset for training and evaluate the model on the same speech and text understanding tasks. Beyond vanilla fine-tuning, we also explore Low-rank Adaptation \citep[LoRA]{hu2021lora} and multitask fine-tuning as they have been shown effective for mitigating catastrophic forgetting in other tasks \citep{xue-etal-2021-mt5, vu2022overcomingcatastrophicforgettingzeroshot}. Details of our fine-tuning setup are shown below:
\begin{itemize}[leftmargin=*, label=\textbullet, itemsep=0pt, topsep=0pt]
    \item \textbf{Vanilla Fine-tuning}: We perform full fine-tuning on the aligned speech-text data with a learning rate of $1 \times 10^{-6}$ and a maximum token size of 4096. Training is model-parallelized across 32 A100 GPUs using Fully Sharded Data Parallel \citep[FSDP]{fsdp}.
    
    \item \textbf{LoRA Fine-tuning}: We leverage the low-rank constraints from  as regularization to prevent model overfitting in MLS dataset. We configure LoRA layers with $\alpha=512$, $r=256$, and a dropout probability of 0.1.
    
    \item \textbf{Multitask Fine-tuning}: To preserve the LLM's pre-trained text capabilities, we also fine-tune SpeechLM on text-only data using Negative Log-Likelihood (NLL) loss. The dataloader is configured to sample from both speech-text and text-only datasets with equal probability. We use the MLS dataset for speech-text training and employ a subset of the \textsc{LLaMa2} training datasets \citep{touvron2023llama} for text-only training.
\end{itemize}

\subsection{Comparison of Fine-tuning Methods}
In \cref{figure::trade_off}, we compare different fine-tuning methods on StoryCloze ($S$) and MMLU. StoryCloze performance is indicative of how well model is fitted to the speech modality and MMLU measures the degree of catastrophic forgetting in pre-trained text abilities. We observe that Vanilla Fine-tuning quickly overfits to the speech domain, achieving improved performance on StoryCloze but drastically decreasing MMLU accuracy. In contrast, LoRA Fine-tuning introduces strong regularization, resulting in limited improvements in speech understanding. Although LoRA mitigates catastrophic forgetting to some extent compared to vanilla fine-tuning, performance still steadily declines. \textbf{Multitask fine-tuning emerges as the most promising approach}, enhancing speech understanding while largely mitigating catastrophic forgetting, evidenced by the modest 2-point drop in MMLU.

Since model performance does not further improve with additional training steps (as shown in \cref{figure::trade_off}), we utilize the checkpoint trained for 5,000 updates to compare with baseline models. The results are presented in \cref{table::finetune_results}. Note that even with only 5,000 updates, the model has observed all speech-text data due to our large effective batch size. As observed from the results, fine-tuned SpeechLM outperforms baseline methods on tasks primarily relying on speech-only information (sWUGGY, sBLIMP, ASR). However, we also observe a decline in performance on $S\rightarrow T$ tasks such as Speech-MMLU and StoryCloze, indicating that \textbf{there is still unavoidable degradation of text capabilities} which adversely affects SpeechLM's cross-modal performance.

Overall, Stage 2 fine-tuning experiments highlight a trade-off between enhanced speech understanding and degraded text abilities when unfreezing pre-trained LLM weights. Though such forgetting phenomenon is unavoidable, our two-stage training pipeline has largely preserved SpeechLM's text ability and our experimental results underscore the importance of incorporating high-quality text data during cross-modal fine-tuning to balance performance across both modalities.

\section{Conclusion}
We propose \method to inject speech representation into pre-trained LLMs. Through explicitly leveraging speech-text alignment, our connector compresses long and sparse speech information to the same granularity as text tokens. With our proposed two-stage training pipeline for modality fusion, \method-based SpeechLM achieves better speech understanding while retaining its pre-trained text ability.

\section*{Limitations}

While our proposed \method significantly enhances speech-text modality fusion and mitigates catastrophic forgetting, there remain several limitations that warrant further exploration.

First, our work focuses on aligning speech semantics with text in large language models (LLMs). While our experiments show that paralinguistic information, such as speech styles, can be preserved and leveraged through in-context learning, we do not explicitly model these aspects. Future work could better encode prosody, speaker identity, and emotional cues to enhance expressive speech generation and nuanced speech understanding.

Second, our experiments on mitigating catastrophic forgetting are conducted primarily on a single language family, using \llamathree \cite{grattafiori2024llama3herdmodels} as the base LLM and \textsc{DinoSR} \cite{dinosr} as the speech encoder. The extent of our method's effectiveness across different architectures and speech encoders remains unverified.

Finally, while our evaluation covers a range of speech and multimodal benchmarks, additional real-world settings, such as conversational speech, noisy environments, and multilingual scenarios, remain unexplored. Extending our methodology to such conditions will be essential for deploying robust, generalizable SpeechLMs.

% \clearpage
\bibliography{custom}

\newpage
\onecolumn
\appendix
\section*{\LARGE{Supplementary Material}}\label{sec:appendix}

\begin{table}[ht]
    \centering
    \footnotesize
    \begin{tabular}{cl}
    \textbf{Appendix Sections}    & \textbf{Contents}  \\ \toprule
    \autoref{sec::aligner} &  \begin{tabular}[c]{@{}l@{}} Specch-Text Aligner Comparison \end{tabular} \\ \midrule
     \autoref{app::paralinguistic} &  \begin{tabular}[c]{@{}l@{}} Non-semantic Information in \method \end{tabular} \\ \midrule
    \autoref{app::dataset_details}     &  \begin{tabular}[c]{@{}l@{}} Dataset Details \end{tabular} \\ \midrule
     \autoref{app::eval}     &  \begin{tabular}[c]{@{}l@{}} Evaluation Details \end{tabular} \\
      \bottomrule
\end{tabular}    
\end{table}

\section{Speech-text Aligners}\label{sec::aligner}
In this section, we provide more details for the aligners that we experimented with to compute segmentation for \method. To summarize, we tried UnitY2 aligner \cite{communication2023seamless}, CTC-based \cite{ctc} aligner (both character-level and subword-level), and CIF-based \cite{dong2020cif} segmentation. We also compare their performance in this section and show that \unity and \charctc aligner work the best; therefore we adopted them in all our experiments presented in the main paper.

\subsection{Aligner Description}
\paragraph{UnitY2 Aligner} The UnitY2 aligner \citep{communication2023seamless} is a forced aligner that computes speech-text alignment using discrete speech units and character-level text tokens. The speech units are derived by applying K-Means clustering to the XLS-R model \citep{xlsr}. The aligner is trained jointly with a non-autoregressive text-to-unit (T2U) model, adopting the architecture of the RAD-TTS model \citep{shih2021radtts} but replacing the target mel-spectrogram with speech units. It first computes a soft-alignment $\softalignment \in \mathbb{R}^{V \times U}$ between the characters and units:
\begin{align}
\textrm{D}_{i,j} & = ||s^{\text{char}}_{i} - s^{\text{unit}}_{j}||_{2}, \\
\softalignment_{i,j} & = \frac{e^{-\textrm{D}_{i,j}}}{\sum_{k}{e^{-\textrm{D}_{k,j}}}} + \textrm{P}_{\text{prior}}(i|j),
\end{align}
where ${\bf s}^{\text{char}}$ and ${\bf s}^{\text{unit}}$ are the outputs of the character and unit encoders, respectively (both encoders consist of an embedding layer and a 1D convolution layer). $\textrm{D} \in \mathbb{R}^{V \times U}$ is a distance matrix with 
$V$ and $U$ representing the vocabulary sizes of characters and speech units. $\textrm{P}_{\text{prior}} \in \mathbb{R}^{V \times U}$ is the Beta-binomial alignment prior matrix to encourage near-diagonal paths \citep{shih2021radtts}. After soft-alignment is computed, the monotonic alignment search (MAS) algorithm~\cite{glow_tts} is applied to extract the most probable monotonic alignment path. 

\paragraph{CTC-based Aligner} 
Since the UnitY2 aligner requires both speech and transcription, it does not support streamable alignment extraction. To enable textless alignment computation, we explored two CTC-based \citep{ctc} aligners. Given the speech features $\bm x$ and text sequences $\bm y$, CTC computes $P(\bm y|\bm x)$ by summing over all valid alignment paths:
\begin{equation}
    P(\bm{y}|\bm{x}) = \sum_{\pi \in \mathcal{B}^{-1}(\bm{y})} P(\pi|\bm{x})
\end{equation}
Here, $\pi$ denotes a possible alignment path that maps to the target sequence $\bm{y}$, and $\mathcal{B}^{-1}(\bm{y})$ represents the set of all valid paths that collapse to $\bm{y}$ after removing blanks and repeated labels. We investigated two CTC variants: one using character-level text sequences (\charctc) and another using subword token sequences (\subctc), which shares the same vocabulary as the LLM model.

\paragraph{CIF-based Speech Connector} For both CTC and UnitY2 aligners, we extract segmentations from the alignments and then apply selection-based compression \cite{tan2024streaming}. We also experimented with Continuous Integrate-and-Fire \citep[CIF]{dong2020cif} as the connector, which is designed to learn segmentation and perform compression simultaneously. Instead of relying on a fixed, pre-computed segmentation, CIF dynamically segments and aggregates speech features by scoring each feature and computing a weighted average. For more details, we refer readers to the paper \citep{dong2020cif}.

\subsection{Aligner Performance Comparison}\label{sec::aligner_compare}

To compare the quality of different aligners, we trained several \method based on different aligners via distillation. We evaluated the aligners using the Librispeech clean test set by computing the Cosine Similarity (\textbf{Cos(\%)}) and Mean Squared Error (\textbf{MSE}) between the compressed representations and text embeddings. Additionally, we performed zero-shot and five-shot ASR with the learned connector. Note that we never explicitly trained the model for ASR tasks, and the base LLM remained frozen during Stage 1 training. Therefore, the model achieves low word error rates (\textbf{WER}) only when the distilled speech representations closely resemble the text embeddings. As shown in \autoref{table::aligner_compare}, the \textsc{UnitY2} aligner brings the speech representations close to their corresponding text embeddings, achieving very low WER in both zero-shot and five-shot ASR settings. Among textless aligners, we found that \charctc performs the best, likely because it has a much smaller vocabulary compared to \subctc, making it easier to learn. Lastly, CIF resulted in suboptimal performance, due to its less accurate alignment, as its segmentation is predicted by accumulating scores without exploiting the monotonicity between speech and text.

\begin{figure*}[t]
    \includegraphics[scale=0.4]{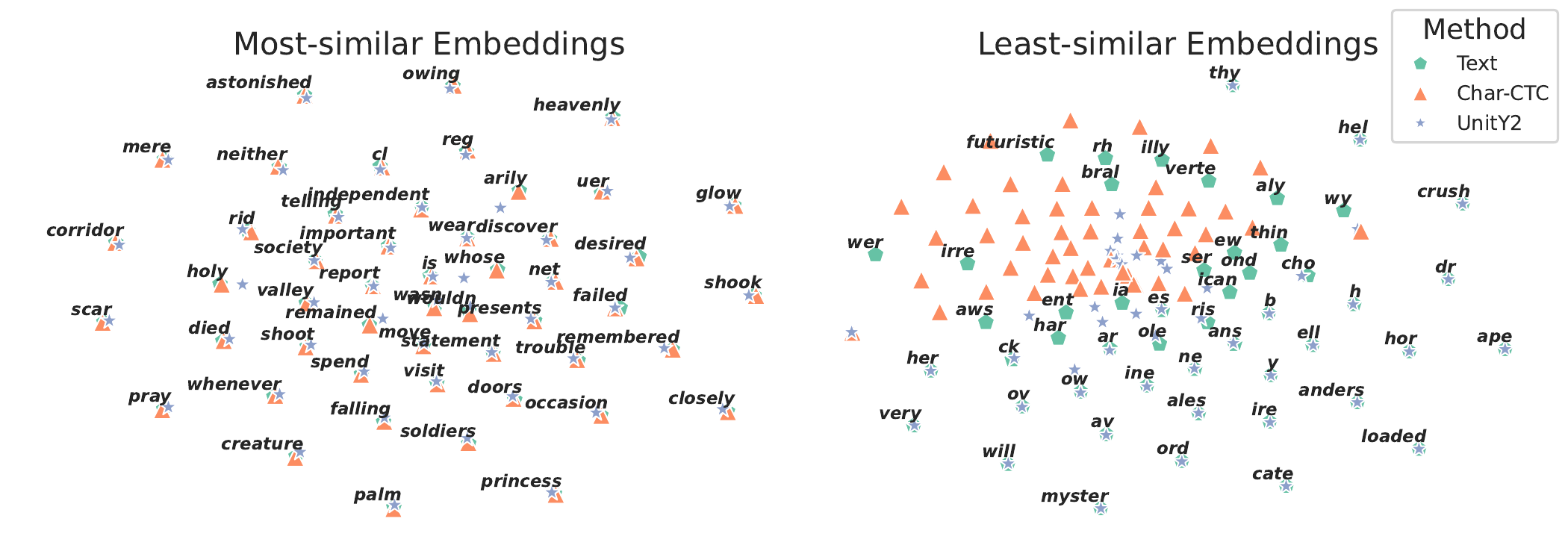}
    % \vspace{-1em}
    \caption{t-SNE plots of text and speech representations after distillation.}
    \label{figure::visualize_embed}
    \vspace{-1em}
\end{figure*}

\begin{wraptable}{r}{0.5\textwidth} % 'r' for right placement, adjust width as needed
    \centering
    \small
    \begin{tabular}{
        l
        S[table-format=2.1]
        S[table-format=1.3]
        c
    }
    \toprule
    \textbf{Model Type} & \textbf{Cos(\%)}$\uparrow$ & \textbf{MSE}$\downarrow$ & \textbf{WER (\%)} $\downarrow$ \\
    \midrule
    \unity         & \textbf{96.8} & \textbf{0.018} & \textbf{5.6 / 4.0} \\
    \charctc       & 95.1 & 0.023 & 9.7 / 6.5 \\
    \subctc        & 92.2 & 0.037 & 16.7 / 14.0 \\
    CIF            & 77.5 & 0.096 & 27.6 / 23.7 \\
    \bottomrule
    \end{tabular}
    \vspace{-1em}
    \caption{Performance comparison (with Cosine Similarity, MSE, and 0/5-shot ASR WER) between different aligners used for Stage 1 training, evaluated on Librispeech.}
    \vspace{-1em} % Adjust vertical space as needed
    \label{table::aligner_compare}
\end{wraptable}

To visualize the effect of distillation, we present t-SNE plots of the adapted speech representations and text embeddings in \cref{figure::visualize_embed}, categorizing them into high and low similarity groups based on the cosine similarity between \charctc representations and text embeddings. We observe that longer subwords tend to exhibit higher similarity, likely because their long segments make it easier for the connector to convert speech representations into corresponding text embeddings. Furthermore, longer subwords possess more coherent semantics compared to shorter tokens. like `wy' or `ia'.

\begin{wraptable}{l}{0.4\textwidth} % 'r' for right placement, adjust width as needed
    \centering
    \small
    \begin{tabular}{lcc}
        \toprule
        \textbf{Aligner} & \textbf{WBE}$\downarrow$ & \textbf{WDUR} \\
        \midrule
        Groundtruth       & 0   & 305 \\
        \unity            & 33  & 279 \\
        \charctc          & 42  & 230 \\
        \midrule
        \textit{\,\,Other Aligners} \\
        CTC+Label Prior   & 29 & 288 \\
        MMS                & 37 & 242 \\
        MFA                & 23 & 314 \\
        \bottomrule
    \end{tabular}
    \caption{Alignment quality comparison.}
    \vspace{-1em} % Adjust vertical space as needed
    \label{table::alignment_quality}
\end{wraptable}

Given that \unity and \charctc performs the best, we also follow \citet{huang2024peakyaccuratectcforced} to measure their word boundary error (WBE) and word average duration (WDUR) using the TIMIT \citep{garofolo1993timit} data. Though the aligner quality can be further improved with other methods such as CTC + Label Prior \citep{huang2024peakyaccuratectcforced}, MMS \citep{mms}, or MFA \citep{McAuliffe2017MontrealFA}, \charctc and \unity still achieve good quality and we choose them out of simplicity and general availability (unlike "CTC+Label Prior", for example, which requires customization with library like k2\footnote{\url{https://github.com/k2-fsa/k2}}).

\section{Beyond Semantics: Speech Style Recognition with In-context Learning}\label{app::paralinguistic}

To explore the non-semantic capabilities of our SpeechLM, particularly its ability to retain and utilize paralinguistic information, we conducted additional experiments focusing on speech style recognition through in-context learning. Specifically, we investigated whether the \method-based SpeechLM (based on the UnitY2 aligner), can differentiate between various speech styles without explicit training on paralinguistic cues.

We utilized the Expresso dataset \cite{expresso_benchmark}, which comprises speeches delivered in distinct styles such as happy, sad, whispering, and laughing. Two primary tasks were designed to assess the model's performance:

\begin{enumerate}
    \item \textbf{Whisper vs. Laugh}: The model was tasked with identifying whether a given speech was whispered or laughed. The prompt provided to the model was:
    \begin{quote}
        "You are given speeches from two styles. Your task is to judge if the speech is a whisper or laugh. Here are some example speeches: [Speech]: \{speech\} [Style]: \{whisper/laugh\}..."
    \end{quote}
    
    \item \textbf{Happy vs. Sad}: The model was asked to determine if the speech was delivered happily or sadly. The prompt used was:
    \begin{quote}
        "Listen to the following speech and judge if the speaker is happy or sad. Here are some examples: [Speech]: \{speech\} [Emotion]: \{happy/sad\}..."
    \end{quote}
\end{enumerate}

For each task, we evaluated the model's performance using varying numbers of in-context examples: 0-shot, 1-shot, 5-shot, and 10-shot. The results, averaged over 10 runs, are presented in Table \ref{tab:style_recognition}. Additionally, we benchmarked a cascaded system comprising Whisper and Llama3 for comparison (this cascaded baseline does no preserve non-semantic information and can only infer the speech style through transcripted content).

\begin{table}[h]
    \centering
    \small
    \begin{tabular}{lccccc}
        \toprule
        \textbf{Task} & \textbf{Model} & \textbf{0-shot} & \textbf{1-shot} & \textbf{5-shot} & \textbf{10-shot} \\
        \midrule
        \multirow{2}{*}{Whisper vs. Laugh} & Cascaded System & 51.6 & 52.1 & 52.2 & 54.7 \\
                                         & Ours            & 49.6 & 62.4 & 64.0 & 75.9 \\
        \midrule
        \multirow{2}{*}{Happy vs. Sad}     & Cascaded System & 50.0 & 51.4 & 51.8 & 51.0 \\
                                         & Ours            & 51.6 & 52.1 & 52.2 & 54.7 \\
        \bottomrule
    \end{tabular}
    \caption{Accuracy of Speech Style Recognition Tasks with In-context Learning}
    \label{tab:style_recognition}
\end{table}

The results indicate that with zero-shot prompting, our model generates predictions close to random chance, as it has not been trained to utilize paralinguistic information. However, with the introduction of a few-shot learning approach, the model significantly improves its ability to distinguish between whispering and laughing speech, achieving up to 75.9\% accuracy with 10-shot examples. This suggests that the model's representations inherently contain paralinguistic information that can be harnessed through in-context learning. For the Happy vs. Sad task, the improvement is modest, peaking at 54.7\% accuracy with 10-shot examples. This lesser performance compared to the Whisper vs. Laugh task may be attributed to the subtler differences in emotional expression compared to the more pronounced style differences between whispering and laughing.

Overall, these findings demonstrate that \textbf{our SpeechLM can effectively leverage in-context learning to recognize different speech styles}, thereby highlighting the presence of paralinguistic information within the model's representations. This capability complements existing methods that incorporate paralinguistic information, such as the use of expressive tokens in SpiritLM \cite{nguyen2024spiritlm} or emotion-relevant instruction tuning in SALMONN \cite{tang2024salmonn}.

\newpage
\section{Datasets}\label{app::dataset_details}

\begin{table*}[h]
\centering
\small
\begin{tabular}{l l l l}
\toprule
\textbf{Eval Dataset} & \textbf{Type} & \textbf{Eval Metric} &\textbf{Eval Modality}\\
\midrule
sWUGGY \citep{nguyen2020zeroresourcespeechbenchmark} & Choice Task     & Accuracy   &$S, T$      \\
sBLIMP \citep{nguyen2020zeroresourcespeechbenchmark} & Choice Task     & Accuracy   &$S, T$      \\
StoryCloze \citep{storycloze_dataset}                & Choice Task     & Accuracy   &$S, T$, $S\rightarrow T$      \\
MMLU \citep{mmlu_dataset}                            & Choice Task     & Accuracy    &$T$     \\
Speech-MMLU (\textit{Ours})                                          & Choice Task     & Accuracy    &$S\rightarrow T$     \\
LibriSpeech \citep{librisppech_dataset}          & Generation Task & Word Error Rate &$S\rightarrow T$  \\
\bottomrule
\end{tabular}
\caption{Evaluation Datasets and their types. For the evaluation format, $S$ is speech-only, $T$ is text-only, and $S\rightarrow T$ means the evaluation prompt consists of speech prefix and text continuation.}
\vspace{-1em}
\label{table::datasets}
\end{table*}

As described in \cref{sec::dataset}, we employ sWUGGY, sBLIMP, StoryCloze, MMLU, Speech-MMLU and Librispeech datasets to assess model performance. In this section, we provide more examples for each evaluation set. sWUGGY and sBLIMP are simple tasks where two choices can be directly compared. As shown in \cref{table::dataset_examples}, sWUGGY provides two choices that require models to discriminate real words from non-words. sBLIMP assesses whether the model can distinguish between a grammatically correct sentence and its ungrammatical variant. 

MMLU and StoryCloze, on the other hand, have a prefix and choices. The StoryCloze dataset measures whether the model can identify the logical ending between two sentences given at the beginning of a short story. Since StoryCloze has a shared prefix, we can synthesize only the prefix part into speech and keep choices in text format, resulting in our $S\rightarrow T$ format evaluation that assess the model's cross-modal understanding. Similarly, for MMLU, we also synthesize its prefix (the question portion) into speech and keep the choices in text format, resulting in our Speech-MMLU dataset. Since some topics have bad audio synthesis quality (\eg the algebra subset contains many mathematical notations), we only keep 22 topics in our test suite (as shown in the ``Topic'' column of \cref{table::speech_mmlu_details}).

\begin{table}[h]
\centering
% \vspace{-1em}
\small
\begin{tabularx}{\textwidth}{@{} l| p{5cm} | p{5cm} @{}}
\toprule
\textbf{Name} & \textbf{Prefix} & \textbf{Choices} \\
\midrule
sWUGGY & N/A & \{Good=obsolete, Bad=odsolete\} \\
\addlinespace
\midrule
sBLIMP & N/A & \{Good=Walter was harming himself, Bad=Walter was harming itself\} \\
\addlinespace
\midrule
StoryCloze & I had been giving this homeless man change every day. He was on the same corner near my house. One day, as I was driving through my neighborhood I saw a new car. Soon enough, I saw the same homeless man emerge from it! & \{Good=I never gave the man money again. Bad=The next day I gave the man twenty dollars.\} \\
\addlinespace
\midrule
MMLU & During the period when life is believed to have begun, the atmosphere on primitive Earth contained abundant amounts of all the following gases except & \{"A": "oxygen", "B": "hydrogen", "C": "ammonia", "D": "methane"\} \\
\bottomrule
\end{tabularx}
\vspace{-1em}
\caption{Examples of different evaluation datasets.}
\vspace{-1em}
\label{table::dataset_examples}
\end{table}

\newpage
\section{Evaluation Metric and Prompt}\label{app::eval}
% \begin{wrapfigure}{r}{0.4\textwidth}
% \vspace{-1em}
% \vspace{-1em}
% \end{wrapfigure}
Choice tasks (sWUGGY, sBLIMP, StoryCloze, MMLU, Speech-MMLU) are evaluated by comparing perplexity of different choices. The choice with smallest perplexity is selected as the prediction and we measure accuracy across different benchmarks.

For generation task (prompt-based ASR), we use the prompt below, with pairs of speech and transcription is provided to the SpeechLM. For 0-shot evaluation, we do not include any examplers.

\begin{tcolorbox}[colback=gray!5,colframe=gray!80,title=Prompt,width=0.5\textwidth]
\textcolor{brown}{
Given the speech, provide its transcription.\\
{[}speech{]}: \{demo speech\}\\
{[}text{]}: \{demo transcription\}\\
...\\
{[}speech{]}: \{speech to transcribe\}\\
{[}text{]}:
}
\end{tcolorbox}

\paragraph{Speech MMLU Evaluation}\label{app::speech_mmlu} We craft speech MMLU by synthesizing the questions of MMLU into audio through \textsc{AudioBox}. Since some domains have bad synthesis quality (such as algebra, which includes many math notations), we filtered those domains out from our evaluation.

We present the detailed comparison results in \cref{table::speech_mmlu_details} for a better comparison of model performance across different domains/topics. We see that the trend for different domains is mostly consistent, with our alignment-aware connector based on \unity achieving the best performance, followed by \charctc based connector. Similar as our main findings, the unit-based system has worse performance due to information loss from discretization and the fine-tuned model suffers from catastrophic forgetting (albeit mitigated through our multitask fine-tuning approach). Nevertheless, all these \method based system obtains better performance compared to \spiritlm (\llamathree), confirming the effectiveness of our modality-fusion strategy.

\begin{table*}[h]
\centering
\small
\resizebox{\textwidth}{!}{
\begin{tabular}{lcccccccccccc}
\toprule
\textbf{Topic} & \multicolumn{2}{c}{\textbf{\spiritlm{} }} & \multicolumn{2}{c}{\textbf{\unity{} + Mask}} & \multicolumn{2}{c}{\textbf{\unity{}}} & \multicolumn{2}{c}{\textbf{\charctc{}}} & \multicolumn{2}{c}{\textbf{Unit-based}} & \multicolumn{2}{c}{\textbf{Fine-tuned}} \\
\cmidrule(lr){2-3} \cmidrule(lr){4-5} \cmidrule(lr){6-7} \cmidrule(lr){8-9} \cmidrule(lr){10-11} \cmidrule(lr){12-13}
 & \textbf{0-shot} & \textbf{5-shot} & \textbf{0-shot} & \textbf{5-shot} & \textbf{0-shot} & \textbf{5-shot} & \textbf{0-shot} & \textbf{5-shot} & \textbf{0-shot} & \textbf{5-shot} & \textbf{0-shot} & \textbf{5-shot} \\
\midrule
Astronomy & 45.6 & 40.8 & 60.0 & 66.0 & 60.7 & 65.3 & 57.0 & 60.4 & 49.7 & 61.1 & 50.7 & 52.0 \\
Business Ethics & 37.1 & 40.2 & 52.0 & 60.0 & 53.0 & 62.0 & 56.0 & 59.0 & 52.0 & 55.0 & 37.0 & 51.0 \\
Clinical Knowledge & 36.0 & 39.8 & 60.6 & 63.3 & 61.0 & 62.9 & 61.2 & 62.7 & 57.8 & 57.4 & 47.3 & 53.8 \\
College Biology & 36.4 & 33.6 & 65.0 & 69.9 & 62.9 & 68.5 & 57.7 & 59.9 & 54.2 & 57.7 & 40.6 & 44.1 \\
Electrical Engineering & 37.7 & 44.2 & 52.5 & 57.4 & 52.5 & 53.9 & 48.2 & 58.9 & 44.7 & 48.2 & 53.2 & 54.6 \\
High School Biology & 40.8 & 41.2 & 66.0 & 72.2 & 67.6 & 72.2 & 63.3 & 68.2 & 57.1 & 65.6 & 50.5 & 62.5 \\
High School Gov. Pol. & 44.4 & 43.4 & 79.2 & 84.9 & 78.1 & 83.3 & 76.6 & 81.8 & 71.4 & 73.4 & 54.7 & 64.1 \\
International Law & 55.9 & 58.5 & 71.1 & 81.0 & 71.1 & 81.0 & 71.1 & 80.2 & 71.1 & 75.2 & 66.1 & 71.1 \\
Jurisprudence & 37.1 & 36.2 & 60.2 & 68.5 & 62.0 & 70.4 & 57.4 & 63.9 & 54.6 & 60.2 & 51.9 & 57.4 \\
Machine Learning & 39.3 & 32.1 & 45.8 & 59.3 & 50.8 & 59.3 & 45.8 & 61.0 & 44.1 & 57.6 & 39.0 & 55.9 \\
Management & 43.0 & 42.0 & 79.6 & 84.5 & 77.7 & 75.7 & 73.8 & 74.8 & 68.0 & 70.9 & 45.6 & 65.0 \\
Marketing & 39.8 & 49.8 & 77.8 & 85.0 & 76.1 & 81.6 & 76.9 & 81.6 & 74.4 & 76.9 & 51.3 & 67.1 \\
Miscellaneous & 38.5 & 36.4 & 69.2 & 71.5 & 66.6 & 70.1 & 60.3 & 64.6 & 52.3 & 57.5 & 42.7 & 50.3 \\
Moral Disputes & 39.1 & 42.3 & 59.5 & 66.5 & 59.5 & 67.3 & 56.4 & 62.7 & 52.9 & 62.1 & 43.6 & 52.9 \\
Nutrition & 45.0 & 47.3 & 68.4 & 69.1 & 66.1 & 66.8 & 65.5 & 62.8 & 64.5 & 59.8 & 52.8 & 58.5 \\
Philosophy & 37.5 & 37.2 & 58.3 & 64.5 & 59.0 & 62.5 & 55.9 & 64.1 & 54.6 & 59.5 & 44.0 & 53.1 \\
Prehistory & 38.9 & 43.3 & 62.0 & 66.4 & 61.1 & 64.5 & 61.2 & 64.3 & 55.0 & 57.5 & 49.1 & 55.2 \\
Security Studies & 43.8 & 54.8 & 63.8 & 67.8 & 61.7 & 67.8 & 68.1 & 76.9 & 59.3 & 69.2 & 51.0 & 59.7 \\
Sociology & 37.4 & 45.5 & 71.6 & 74.6 & 68.7 & 74.6 & 69.7 & 73.6 & 68.2 & 72.1 & 57.7 & 66.2 \\
US Foreign Policy & 56.7 & 60.8 & 80.0 & 80.0 & 78.0 & 85.0 & 75.8 & 81.8 & 75.8 & 83.8 & 61.0 & 76.0 \\
Virology & 40.1 & 46.3 & 47.9 & 49.1 & 49.1 & 53.9 & 47.9 & 49.7 & 46.1 & 51.5 & 46.7 & 44.8 \\
World Religions & 39.3 & 46.4 & 66.1 & 67.8 & 63.2 & 63.7 & 52.0 & 59.1 & 51.5 & 60.8 & 40.9 & 50.3 \\
Micro Average & 40.5 & 42.7 & 65.0 & 69.5 & 64.2 & 68.6 & 61.7 & 66.5 & 58.1 & 63.3 & 49.0 & 57.5 \\
\bottomrule
\end{tabular}}
\caption{Detailed Speech-MMLU evaluation results on different domains.}
\label{table::speech_mmlu_details}
\end{table*}

\end{document}